# RECITYGEN

*Interactive and Generative Participatory Urban Design Tool with Latent Diffusion and Segment Anything*

Di Mo †
Peking University
modi2022@stu.pku.edu.cn

Mingyang Sun †
Massachusetts Institute of Technology
msun14@mit.edu

Runjia Tian
Harvard University
runjiatian@gmail.com

Chengxiu Yin
Columbia University
cy2757@columbia.edu

Yanhong Wu
Peking University
ywuyh@pku.edu.cn

Liyan Xu*
Peking University
xuliyan@pku.edu.cn

**Abstract**. Urban design profoundly impacts public spaces and community engagement. Traditional top-down methods often overlook public input, creating a gap in design aspirations and reality. Recent advancements in digital tools, like City Information Modelling and augmented reality, have enabled a more participatory process involving more stakeholders in urban design. Further, deep learning and latent diffusion models have lowered barriers for design generation, providing even more opportunities for participatory urban design. Combining state-of-the-art latent diffusion models with interactive semantic segmentation, we propose RECITYGEN, a novel tool that allows users to interactively create variational street view images of urban environments using text prompts. In a pilot project in Beijing, users employed RECITYGEN to suggest improvements for an ongoing Urban Regeneration project. Despite some limitations, RECITYGEN has shown significant potential in aligning with public preferences, indicating a shift towards more dynamic and inclusive urban planning methods. The source code for the project can be found at RECITYGEN GitHub.

**Keywords.** Urban Design, Urban Renewal, Participatory Design, Artificial Intelligence, Streetview, Mass-Customization, User-Interface

## 1. Introduction

The late 20th century heralded a pivotal shift in urban design, profoundly redefining the concept of 'place'. This era recognized 'place' as a multifaceted entity, deeply rooted in human environmental experiences and spatial cognition, thus centralising it in urban design discourse. This philosophical shift underscored placemaking as a participatory endeavour, significantly emphasising the enhancement of physical urban spaces through community engagement.

This period also witnessed the rise of diverse participatory urban planning models. These models were predicated on the principles of inclusivity and democratic participation, focusing on integrating community members directly into the planning process. This approach ensured that urban spaces were not only reflective of but also responsive to, the local community's needs and aspirations. Consequently, this shift towards participatory urban design has been crucial in fostering more livable, sustainable, and community-centric urban environments. It repositioned streets and placesT not merely as physical entities, but as dynamic spaces shaped by and for the people who inhabit them, emphasising the role of communal input in shaping the urban fabric.

† indicates equal contribution to this work *Corresponding author

## 2. Related Works

### 2.1. PARTICIPATORY URBAN DESIGN

Participatory Urban Design (PUD) emphasizes community engagement in urban planning, blending diverse methods and technologies to address complex issues. Central to PUD are empowerment and building community connections, as highlighted by Sanoff (2008). Technology in PUD, discussed by Vainio (2016), serves as both a tool and a medium for innovation. Wagner et al. (2009) demonstrate how participatory technologies aid community practices, with tools like RECITYGEN enabling collaborative design. Understanding power dynamics, as explored by Bratteteig and Wagner (2012), is essential in participatory decision-making. PUD's role in sustainable development, emphasized by Kee (2019), requires multidisciplinary collaboration. Bina and Ricci (2016) show the importance of participatory scenarios in policy debates.

### 2.2. SEGMENTATION ANALYSIS OF URBAN STREET VIEW

The evolution of deep learning in convolutional neural networks (CNNs) has been particularly impactful in image segmentation and object detection, marking significant advancements from 2012 to 2019 (Hoeser and Kuenzer, 2020; Minaee et al., 2020).In urban planning, deep learning plays a vital role in processing the vast data generated within city environments, aiding in better understanding and improving public services (Chen et al., 2019). Zhang et al. (2021) used deep semantic segmentation with street view images for air pollution prediction. Xia et al. (2021) developed the Panoramic View Green View Index to assess street-level greenery. Chang et al. (2020) linked street view elements to urban land use, while Tao et al. (2022) correlated street semantics with human activity density for urban planning insights. Digital tools like City Information Modelling (CIM) and augmented reality are being used to bridge the gap between urban designers and the public, as discussed by Stojanovski et al. (2020) and Hanzl (2007). Its application extends to participatory urban planning, where it aids in detecting urban informality and enhancing participatory data analytics (Ibrahim et al., 2018; Lock et al., 2021).

### 2.3. URBAN STREET VIEW GENERATION

# RECITYGEN： INTERACTIVE AND GENERATIVE PARTICIPATORY URBAN DESIGN TOOL WITH LATENT DIFFUSION AND SEGMENT ANYTHING

Recent advancements in latent diffusion models and semantic segmentation have further refined deep learning's capabilities in urban design. Clint et al. (2022) and Kwieciński and Slyk (2023) introduced tools for greenery evaluation and apartment layout generation. Kim et al. (2022) proposed PlacemakingAI for real-time urban streetscape generation, highlighting the role of generative models in participatory design. These advancements indicate effective use of deep learning models in urban street view generation and analysis for more sustainable and engaged urban environments. Combined with advanced interactive segmentation algorithms such as Segment Anything Model (SAM) (Kirillov et al., 2022) we see a novel approach for interactive and generative tools for participatory urban design.

## 3. Methodology

RECITYGEN is composed of three important components:

- **Algorithm:** RECITYGEN achieves the state-of-the-art semantics-guided urban street view generation by marrying Segment Anything Model(SAM) and state-of-the-art latent diffusion model Stable Diffusion.
- **Web-Based User Interface:** We developed the RECITYGEN prototype as a Web App and hooked the interface with the algorithm backend using Gradio UI framework. The Web App is easily accessible on users' smartphones and tablets, where anyone can easily identify urban environments that need improvement. In addition, we developed a POI tracker where users can geolabel the urban space marked for improvement, creating a crowd-sourcing platform for continuous feedback of urban environments.
- **Usability Study:** After deploying the backend algorithm and frontend WebApp, we performed an on-site user study to collect feedback about RECITYGEN so that we could continuously improve the workflow. We carried out our pilot testing in an urban community in Beijing where users with diverse backgrounds tested out the app.

### 3.1. CONDITIONAL URBAN STREET VIEW GENERATION WITH SEGMENT ANYTHING AND LATENT DIFFUSION

Within RECITYGEN, we used state-of-the-art deep learning models in both urban street view segmentation and generation. Using the interactive semantic-based segmentation from SAM, we could street the generation of Stable Diffusion and accurately change partial infills of urban street views by connecting the segmentation from SAM to the input of an inpainting-based diffusion model.

**Segment Anything Model (SAM):** SAM introduces an essential element to RECITYGEN, furnishing it with advanced interactive semantic segmentation capabilities. As a leading-edge deep learning model, SAM is engineered for the precise identification and segmentation of diverse elements within urban images. This comprehensive training equips SAM to distinguish accurately between various components of the urban landscape, including buildings, roads, and green spaces. By a simple click on the screen, users can get a segmentation image from an arbitrary urban street view including photos from one’s smartphone.

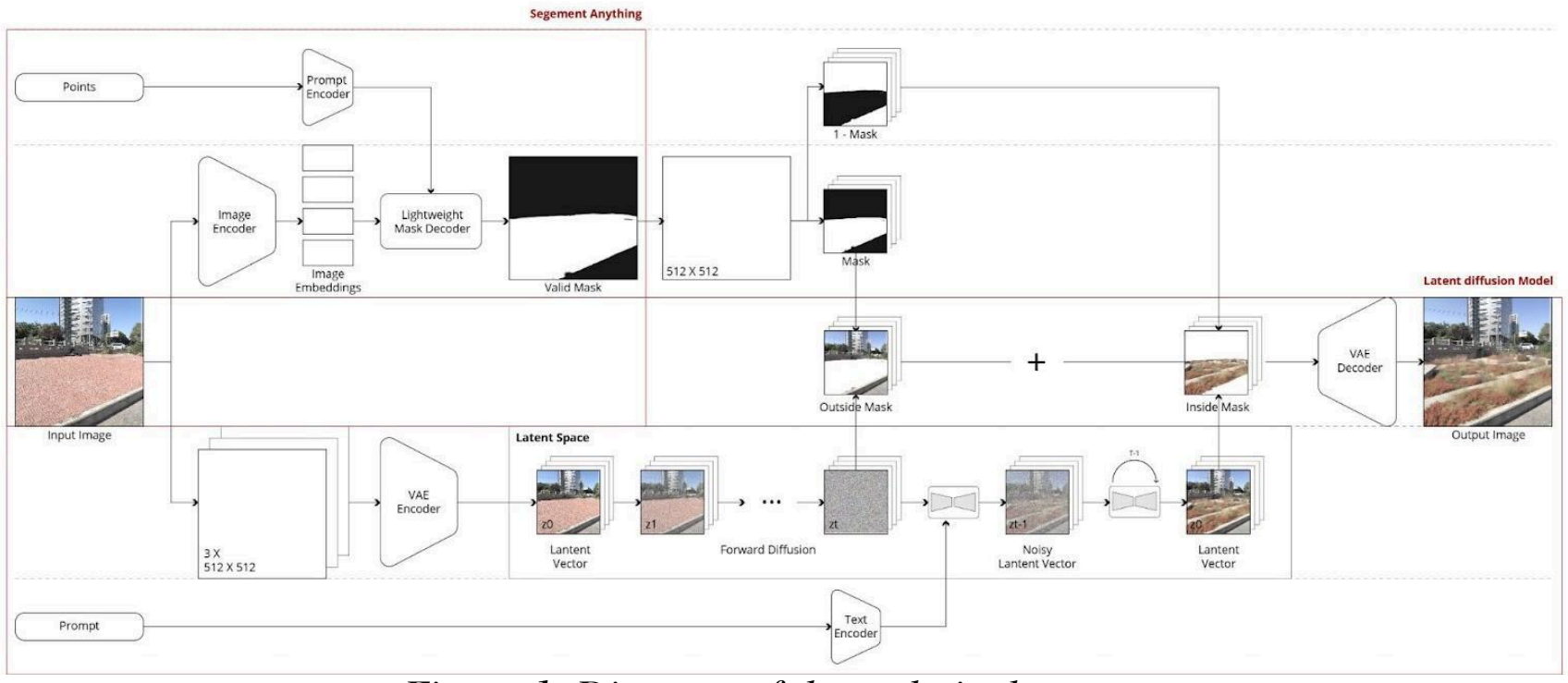

*Figure 1. Diagram of the technical process*

**Latent Diffusion Model:** Inpainting-based Stable Diffusion model stands as another core component of RECITYGEN. It is adept at creating detailed and high quality urban street view images conditioned on the text and alpha mask input. The model operates on a latent diffusion process, wherein it is trained to methodically transform a random noise distribution into a structured image, aligning with specific textual inputs. This transformation unfolds through a series of iterative steps, each enhancing the image's detail and coherence. The model's proficiency lies in its ability to capture and replicate complex urban patterns and textures, making it highly suitable for visualising a wide array of urban environments as described by user inputs. The Stable Diffusion model's primary strength resides in its capacity to interpret various urban scenarios from simple text descriptions, thereby enabling the production of detailed visualisations that align precisely with participant-provided urban design suggestions.

Unlike traditional PUD tools where users have to draw the segmentation themselves, we simplify the user cognitive load to the minimum where users only need to click on the area they would like to modify and input a text description to get an accurate improved urban street view image.

## 3.2. WEB-BASED USER INTERFACE DESIGN AND DEVELOPMENT

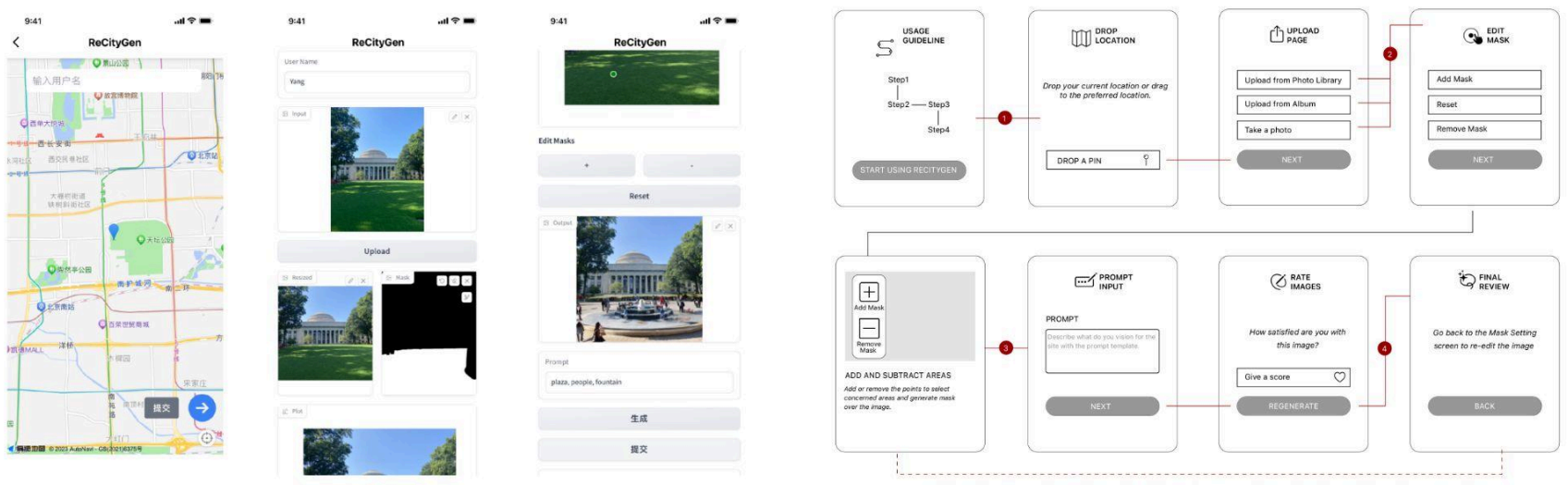

*Figure 2. Information Architecture and User Interface of RECITYGEN*

The front end of RECITYGEN is composed of three major components: geographic data labelling, interactive segmentation and prompt-based image generation.

- **Geolocation labelling**: RECITYGEN is developed to be a crowd-sourcing feedback tool for urban design, thus it is important for users to label the geographical location for the urban environment on which they will provide feedback. After a user enters the GPS location on the map, we generate a feedback ID and use it as an identifier for the feedback entry.
- **Interactive Segmentation Mask Generation**: Once a user creates the feedback entry, the user can move on to the generation step. With Gradio, we implemented a similar user interface as the original SAM interface, where users can click on the screen to generate segmentations from the '+'(include key point) and '-'(exclude key points).
- **Urban Street View Generation**: We use the output from the SAM model as the input for inpaiting-based stable diffusion to guide the generation. In the meantime, users are able to input text prompts as a guide to the content of the generation, making the synthesis of urban street view highly controllable.

## 3.3. USABILITY STUDIES

### 3.3.1. *User Testing*

Our RECITYGEN evaluation involved a diverse participant group, varying in age, gender, technical skill, and urban design knowledge. The test environment simulated real urban settings, using virtual simulations or physical setups, guided by methodologies like those in "The Routledge Handbook of Urban Design Research Methods." The testing unfolded in phases:

- **Initial Interaction.** Participants explored RECITYGEN's interface and features, providing initial feedback on its intuitiveness and user-friendliness.

- **Task-Based Evaluation**: They undertook tasks mimicking real urban design challenges to assess the tool's practical effectiveness.
- **Advanced Interaction Analysis.** This phase examined RECITYGEN's advanced capabilities and adaptability to diverse urban design issues.
- **Feedback Compilation.** Through surveys and interviews, comprehensive feedback on usability, functionality, and applicability was gathered.
- **Data Synthesis and Refinement**. Analysing this feedback, we identified common themes, user experience issues, and potential improvements for RECITYGEN's iterative development.

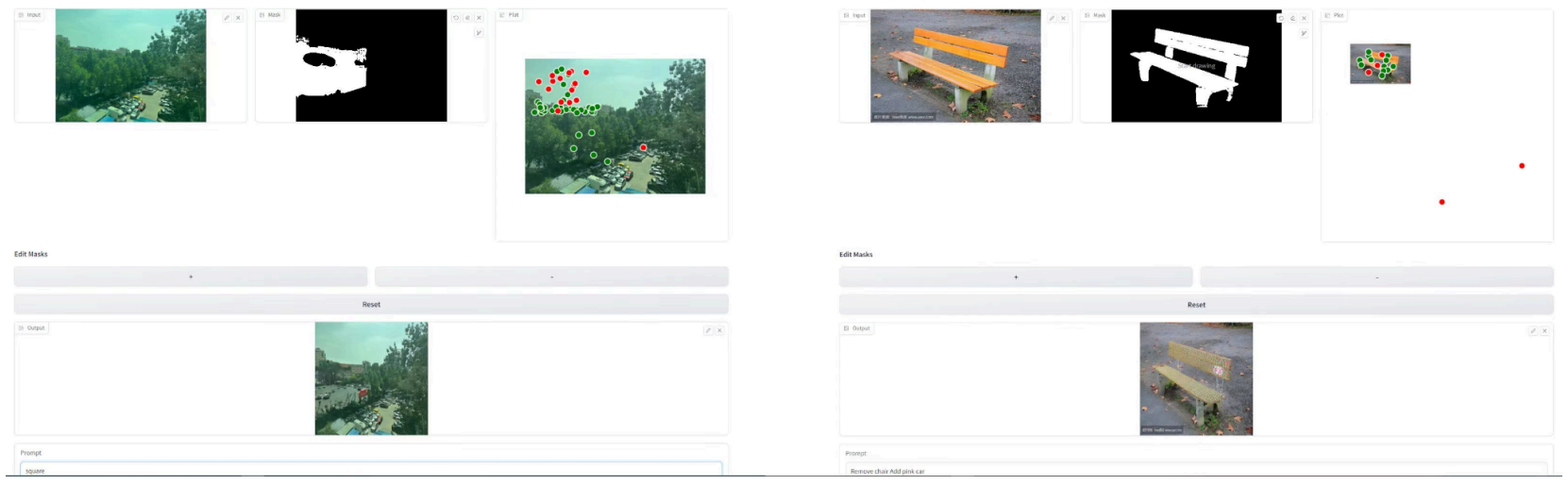

*Figure 3. Web App User Interface in Real World Use Case*

This methodical approach ensured a robust and academically grounded evaluation, aiming to enhance RECITYGEN's real-world urban design applicability.

### 3.3.2. *Questionnaire*

A structured questionnaire was used to capture user feedback, encompassing several key areas as shown in Table 1:

| # | Question | Answer Type |
|---|---|---|
| Section 1: Satisfaction and Continued Use: | | |
| 1 | On a scale from 1 to 5, how would you rate your overall satisfaction with RECITYGEN? | Scale 1 - 5 |
| 2 | How likely are you to continue using RECITYGEN in your future projects | Scale 1 - 5 |
| 3 | On a scale of 1 to 5, how likely are you to recommend RECITYGEN to your peers? | Scale 1 - 5 |
| Section 2: Perceived Effectiveness and User Experience: | | |
| 4 | Please rate your agreement with the statement: 'RECITYGEN rapidly generates the urban scenes I envision | Scale 1 - 5 |
| 5 | Rate how well the images generated by RECITYGEN meet your requirements | Scale 1 - 5 |
| 6 | Express your agreement with the statement: ‘I prefer designing public urban spaces myself over having them designed by experts’ | Scale 1 - 5 |
| 7 | Rate the convenience and user-friendliness of RECITYGEN's operational process (scale 1-5) | Scale 1 - 5 |
| Section 3: Demographic and Background Information | | |
| 8 | Collection of gender, educational level, birth year, profession, and design background. | Text Input |
| Section 4: Open-Ended Feedback and Suggestions | | |
| 9 | Invitation for suggestions on improvements or modifications for RECITYGEN. | Text Input |

*Table 1. Questionnaire Design*

The questionnaire was crafted to ensure precision and avoid ambiguity, employing a Likert scale to facilitate nuanced analysis of user perceptions and experiences.

## 4. **Experiment**

### 4.1. ON-SITE PILOT TESTING

The pilot testing was launched in Beijing's Haidian District, particularly around the Gao Liang Qiao Site, providing real-world insights into how RECITYGEN can transform the PUD process.

In our study's 'Street Design Improvements' section, focusing on RECITYGEN, we observed a notable transformation of a previously underused street corner garden in Beijing's Haidian District. This section of the investigation provides an in-depth analysis of the transformation, supported by user feedback and output data from the tool.

Key initial input parameters such as 'inviting', 'green', and 'community-focused' were effectively interpreted by RECITYGEN into practical urban improvements. These enhancements included the addition of lush greenery, user-friendly seating areas, and enhanced lighting. The effectiveness of RECITYGEN is visually demonstrated through comparative imagery showing the area before and after the intervention.

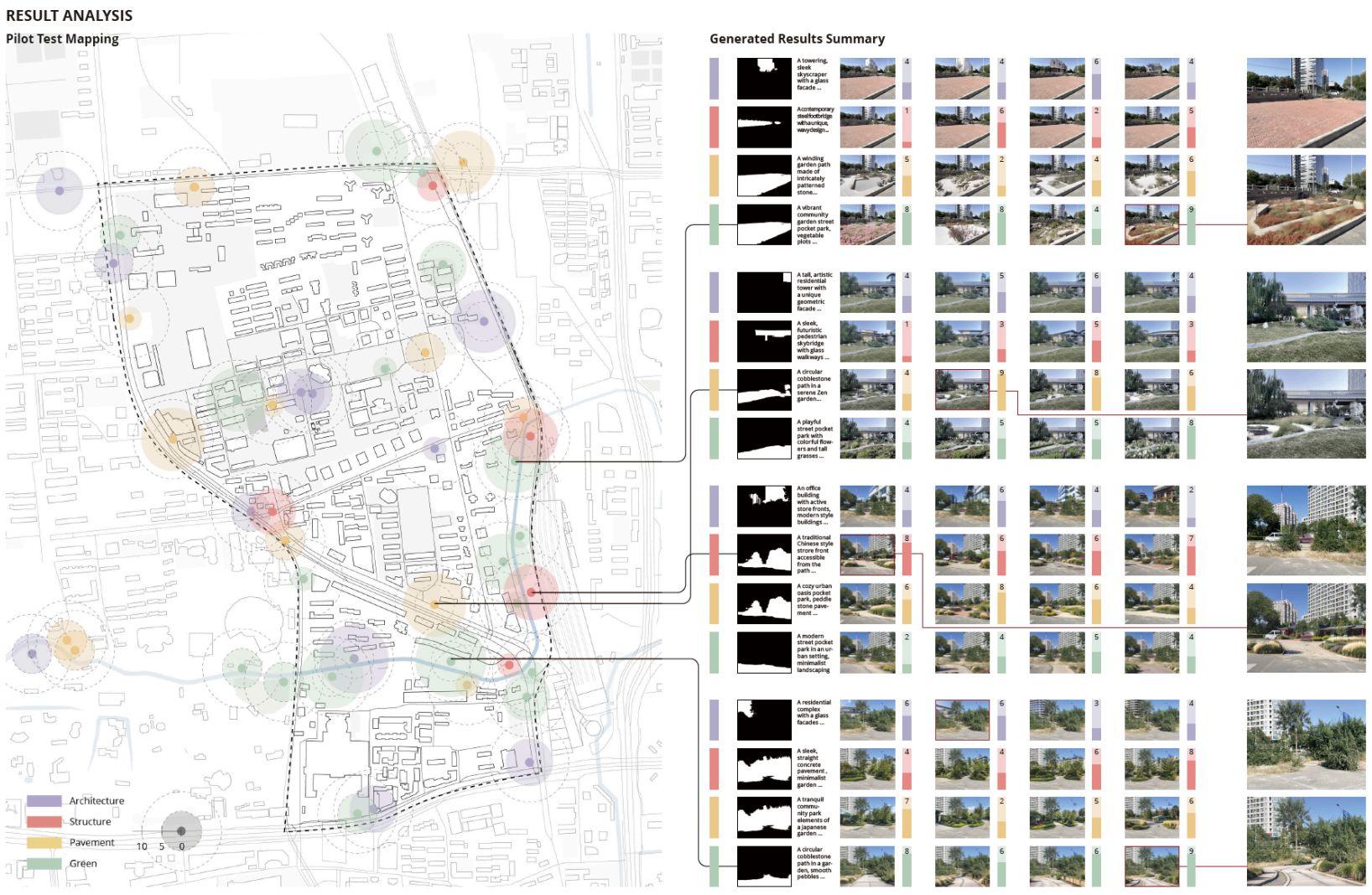

*Figure 4 User Feedback from On-Site Pilot Testing*

## 4.2. QUESTIONNAIRE RESULT ANALYSIS

### 4.2.1 QUANTITATIVE ANALYSIS

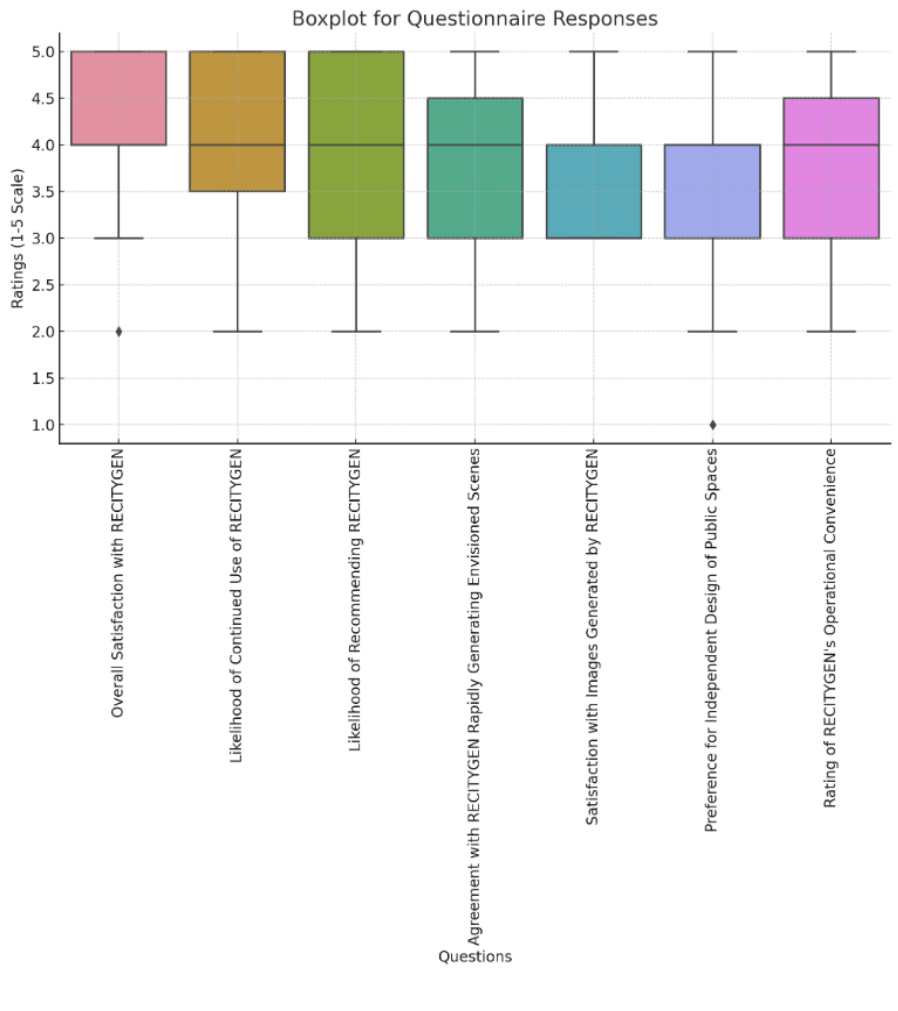

*Figure 6 illustrates the distribution of user satisfaction scores for RECITYGEN.*

In the 'User Engagement' section of our study, we meticulously evaluated how participants interacted with RECITYGEN, focusing on their experiences during the pilot testing in Beijing's Haidian District. Local residents were asked to use the tool to provide ideas in redesigning the street corner garden. Users were then asked to rate the images generated by the tool depicting proposed street renovations using the questionnaire from Section 3.4. This process includes onboarding cycles, where user feedback was utilised to progressively refine and adjust the images. After approximately 3 to 4 iterations of training and onboarding, we observed that users were able to produce images that met their basic satisfaction criteria. This finding underscores the significance of accessibility and ease of use for participatory design tools.

While the majority of participants (9 out of 12) gave high satisfaction ratings (4 or 5), there is a notable minority who expressed moderate satisfaction (scores of 2 and 3), suggesting areas for further improvement. The data indicates a generally positive reception towards RECITYGEN. However, the diversity in satisfaction levels points to potential discrepancies in user expectations or experiences.

### 4.2.2 QUALITATIVE ANALYSIS

From the user experience feedback question, we gathered the following key insights where RECITYGEN can be improved:

**An increased set of 'starter prompts':** Users would frequently express the need for a set of guiding prompts instead of starting from scratch. This shows that although text-based generative models are already intuitive to use, the prompt engineering itself sometimes still has barriers or challenges.

**Need-specific generative models:** For our test case, we mainly used the vanilla Stable Diffusion inpainting model for the generation task. This model could be improved by fine-tuning on urban street view dataset to enhance its a-prior knowledge about urban environment.

**Improvements in the precision of area selection within the tool.** This shows the user's need for an accurate segmentation model to be able to more accurately steer the generation.

**More freedom in post-processing the inpainting mask.** Due to technical limitation, we did not implement inpainting customization for users. However, such features could add more creativity to the workflow and could potentially be more effective for users with a design background.

We also created a word cloud for analysing the prompts from user's Keywords to highlight the most frequently mentioned key words from user's prompt inputs.

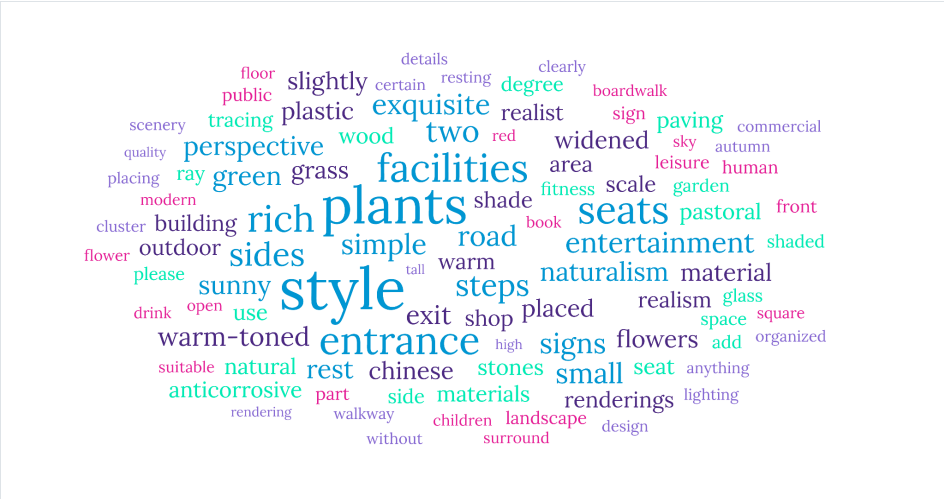

*Figure 5 Prompt Word Cloud*

The feedback we obtained from users lays a solid foundation for future prototyping of RECITYGEN and other AI-powered next generation participatory urban design tools.

## 5. Conclusion

Our involvement in the RECITYGEN research highlighted its effectiveness in community-engaged urban design, as seen in a Beijing case study. While the tool successfully translates user input into urban solutions, it needs improvement in visual detail rendering, especially for complex elements like street furniture and landscaping.

During the pilot phase, we noted RECITYGEN's limitations. While effective in broad conceptualization, it sometimes falls short in depicting realistic details crucial for urban planning. This experience underlined the need for enhancements in its design algorithms and user interface.

We recommend several improvements for RECITYGEN: broadening its range to cover diverse urban contexts and refining user interaction methods, possibly through more intuitive interfaces or augmented reality features. Future research should focus on extensive public engagement and a wider spectrum of user feedback to ensure RECITYGEN meets the evolving needs of urban communities. An iterative development strategy, driven by constant user input and technological updates, is vital for maintaining RECITYGEN's relevance in urban design.